\documentclass[11pt]{article}

\usepackage[]{acl}

\usepackage{times}
\usepackage{latexsym}

\usepackage[T1]{fontenc}

\usepackage[utf8]{inputenc}

\usepackage{microtype}

\usepackage{inconsolata}

\usepackage{graphicx}

\title{Open-Source FUTO Keyboard Layouts and Predictive Text for a Few Endangered Languages}

\author{Mika Hämäläinen\textsuperscript{1}, Jack Rueter\textsuperscript{2} \& Khalid Alnajjar\textsuperscript{3} \\
  \textsuperscript{1}Metropolia University of Applied Sciences\\
  \textsuperscript{2}University of Helsinki \\
  \textsuperscript{3}Rootroo Ltd \\
  \texttt{first.last@metropolia.fi/helsinki.fi/rootroo.com} \\}

\begin{document}
\maketitle
\begin{abstract}
We present open-source FUTO Keyboard layouts and predictive text dictionaries for five endangered Uralic languages: Erzya, Moksha, Komi-Zyrian, Livonian, and Skolt Sami. Existing Divvun Android keyboard layouts are converted automatically into the YAML format used by FUTO Keyboard. For predictive text, we construct weighted AOSP dictionaries using translated Finnish lemma frequencies together with morphological frequencies extracted from small target-language corpora using finite-state transducers. This combination provides a practical way to rank inflected word forms when large monolingual corpora are unavailable. The resulting resources offer reusable mobile keyboard support for endangered Uralic languages.
\end{abstract}

\section{Introduction}

For many endangered and minority languages, digital language support remains limited (see \citealt{joshi-etal-2020-state}. Even when a language has an established orthography and linguistic resources such as dictionaries or morphological analyzers, basic tools for everyday communication may still be missing. A keyboard that provides the correct characters makes it possible to write a language on a phone, but modern users also expect features such as word completion and predictive text.

The Divvun project\footnote{https://divvun.no/en/index.html} has made an important contribution to mobile support for many Uralic and other minority languages by providing keyboard layouts and language technology resources. In this work, we build on these existing resources by converting Divvun keyboard layouts for use with FUTO Keyboard\footnote{https://github.com/futo-org/android-keyboard}, an open-source Android keyboard. Our focus is not only on reproducing the keyboard layouts, but also on adding predictive text resources that can be distributed and used together with them.

Predictive text presents a more difficult problem than keyboard layout conversion. A useful prediction system normally depends on frequency information: common words should be suggested more readily than rare ones, and common inflected forms should be preferred over forms that are grammatically possible but unlikely in actual use  \cite{darragh1991adaptive,gerz2018language}. For well-resourced languages, such frequencies can be estimated from large corpora. This is not straightforward for endangered Uralic languages. Their available corpora may contain enough material to illustrate vocabulary and morphology, but they are often too small to provide reliable frequency estimates for a large predictive dictionary. At the same time, purely rule-based generation does not by itself provide a good way of ranking the many word forms that a morphologically rich language can produce.

We address this problem by combining information from two different sources\footnote{Our code and models are available on GitHub: https://github.com/mikahama/kbdgen2futo}. First, we use lemma frequencies from a large Finnish dataset as a proxy for general lexical frequency. Rather than transferring Finnish word forms directly, we translate frequent Finnish lemmas into the target languages using dictionaries. This provides an approximate ranking of target-language lemmas even where large target-language corpora are unavailable.

Second, we estimate the frequencies of morphological forms from the target languages themselves. Small corpora in each language are analyzed using finite-state transducers (FSTs), allowing us to observe how frequently different morphological analyses and inflectional patterns occur in actual text. We then combine the translated lemma-frequency information with the target-language morphological frequencies to construct weighted predictive text dictionaries. The method is applied to five endangered or minority Uralic languages: Erzya (myv), Moksha (mdf), Komi-Zyrian (kpv), Livonian (liv), and Skolt Sami (sms).

\section{Related Work}

 Mobile interfaces are typically designed around larger languages, which also receive features such as autocorrection, word completion, and predictive text. \citet{lackaff-moner-2016-local} show how this difference can influence language choice in practice: in their study of Irish language users, participants reported difficulties with accented characters as well as frustration caused by autocorrection and predictive features designed for English. Thus, even when the orthography of a minority language can technically be entered using a general-purpose keyboard, the additional effort involved may still favour use of the majority language.

\citet{hamalainen-alnajjar-2021-developing} developed Livonian keyboard layouts for desktop operating systems as well as mobile keyboards for Android and iOS. Because Livonian contains a relatively large number of accented characters, the mobile layout placed many of these characters under their unaccented counterparts rather than assigning each one a dedicated key. Their mobile keyboards also included predictive text based on words from a Finnish--Livonian dictionary. This solution provided lexical coverage despite the scarcity of Livonian data, but prediction was restricted to dictionary lemmas.

Related problems have been addressed for Indigenous languages outside the Uralic family.  \citet{santos-harrigan-2020-design} developed and evaluated a smartphone keyboard for Plains Cree syllabics, combining ergonomic considerations with corpus statistics when designing the layout. More recently, \citet{shames-2025-designing} developed a mobile keyboard for Itunyoso Triqui, where lexical tone and multigraphs make ordinary QWERTY input inconvenient. The resulting layout includes dedicated support for multigraphs and several methods for entering tone. \citet{kakolu-ramarao-2026-keyboards} describes mobile and desktop keyboards for Idu Mishmi, designed to support an orthography containing characters and multi-codepoint Unicode sequences that are difficult to enter with standard keyboards.

Early mobile prediction systems relied primarily on lexicons and word-frequency information. For example, \citet{dunlop-crossan-2000-predictive} studied dictionary-based disambiguation and word completion for mobile-phone input, demonstrating the importance of the lexicon and the ranking of candidate words. More recent systems have employed statistical and neural language models. \citet{yu-etal-2018-device}, for example, developed a neural word-prediction model specifically designed to operate under the memory and latency constraints of a mobile device. Such approaches work well when sufficiently large and representative training data are available, but this assumption is problematic for endangered languages, where even the total available corpus may be small.

Morphologically rich languages pose an additional difficulty because prediction cannot be reduced to selecting among a moderate number of independent word types. A single lemma may occur in many inflected forms, while many theoretically possible forms may be rare or absent in ordinary language use. \citet{lane-bird-2020-interactive} address this problem for Kunwinjku by using a finite-state morphological model for interactive word completion. Instead of enumerating every possible complete word, their method maps partially entered forms to possible completions up to morphological boundaries. The approach was subsequently applied to Plains Cree by  \citet{lane-etal-2022-interactive}, who used an existing FST and investigated different strategies for ranking the resulting morphological completions.

\section{Building FUTO keyboards}

\subsection{Layout conversion}

Both Divvun and FUTO Keyboard define keyboard layouts using YAML files, but the structure and expected fields of the two formats differ. Because of this, Divvun keyboard layouts cannot be used directly in FUTO Keyboard.

Divvun layout files may contain separate configurations for several device and operating system targets, including Android, iPhone, iPad and macOS. Since FUTO Keyboard is an Android keyboard, we use the Android layout definitions as the source for conversion.

To avoid manually rewriting each layout, we implemented a conversion script that reads the Android section of a Divvun YAML file and produces the corresponding FUTO Keyboard YAML format. The script converts the key rows and relevant layout information while adapting the representation to the structure expected by FUTO Keyboard.

This makes it possible to reuse existing Divvun keyboard layouts with relatively little manual work. In this study, the same conversion process was used for the Erzya, Moksha, Komi-Zyrian, Livonian, and Skolt Sami layouts.

\subsection{Building predictive dictionaries}

FUTO Keyboard uses the AOSP dictionary format adopted from HeliBoard for predictive text\footnote{https://codeberg.org/Helium314/aosp-dictionaries}. These dictionaries can associate words with numerical weights, allowing more frequent or more likely words to be ranked higher in the suggestion list. The format can also encode next-word information for context-sensitive prediction.

In this work, we only make use of unigram entries, i.e. individual words together with their weights. We do not include next-word prediction. This keeps the dictionary construction process independent of large target-language corpora, which would be needed to estimate reliable word-to-word transition frequencies. The main task is therefore to generate a sufficiently broad set of word forms and assign useful relative weights to them.

We start by obtaining frequencies for Finnish lemmas in a large online corpus, for this, we use the frequencies provided in SemFi \cite{hamalainen2018extracting}. These frequencies are later mapped to endangered language lemmas using the FST dictionaries in UralicNLP \cite{hamalainen2019uralicnlp}.

Next, we need to know what words there are in each language. We take the FSTs for Erzya, Moksha \cite{rueter2020open}, Skolt Sami \cite{rueter2020fst}, Komi-Zyrian \cite{rueter2021overview} and Livonian \cite{rueter2014livonian} and get lemma lists from their LEXC stem files. Then we proceed to inflect these lemmas to most of their possible inflections using the method and inflection limits described in \citet{hamalainen2021neural}. This way, we end up with large datasets of words and their inflectional forms.

In order to gather language specific understanding of frequencies of word forms, we build a small corpus for each language\footnote{Corpus size in tokens: kpv (5m), liv (3k), mdf (584k), myv (1.2m) and sms (69k) }. We then analyze the words of these corpora using the FST analyzers, the analyzers output morphological readings for each word such as \textit{+N+Pl+Par} and \textit{+V+Act+Imprt+Prs+ConNeg+Sg2}. We discard the actual words and only count the frequencies of these morphological readings.

\subsubsection{Estimating frequencies for inflected word forms}

In this section, we describe how we combine Finnish lemma frequencies with endangered language morphological frequencies. For a word form $w$ with lemma $l$, part of speech $p$, and morphological analysis $m$, we estimate its probability as

\[
P(w) \approx P(p)\,P(l \mid p)\,P(m \mid l,p).
\]

Here, $P(p)$ describes how frequent the relevant part of speech is in the target-language corpus, $P(l \mid p)$ estimates the frequency of the lemma, and $P(m \mid l,p)$ estimates how likely the particular inflection is for that lemma. This decomposition allows lexical frequency to be transferred from Finnish while the distribution of morphological forms remains language-specific.

Finnish frequencies are not used as raw counts in the target-language model. This distinction is important because the Finnish corpus is several orders of magnitude larger than the target-language corpora. Instead, the Finnish counts are converted into a probability distribution over target-language lemmas. If a Finnish lemma has several possible translations, its frequency mass is divided between the corresponding target lemmas rather than being copied in full to every translation. Conversely, if several Finnish lemmas map to the same target-language lemma, their contributions are combined.

We additionally flatten the Finnish frequency distribution using a temperature parameter. If $r_l$ is the Finnish-derived frequency assigned to target lemma $l$, its normalized weight is calculated as

\[
q(l \mid p)
=
\frac{(r_l+\epsilon)^{\tau}}
{\sum_{l'}(r_{l'}+\epsilon)^{\tau}},
\]

where the sum is taken over lemmas belonging to the same part of speech. The small constant $\epsilon$ prevents unmapped words from receiving zero probability, while $0 < \tau \leq 1$ controls how strongly Finnish frequency differences are preserved. Values below one reduce the influence of very large frequency differences in Finnish. This is desirable because frequency in Finnish is informative about general lexical commonness, but it should not be assumed to reproduce the lexical distribution of another language exactly.

When the target-language corpus itself contains occurrences of a lemma, these observations are combined with the Finnish-derived distribution. Finnish frequencies are treated as a prior rather than as additional corpus tokens. In practice, a configurable pseudo-count parameter determines how much influence the Finnish distribution has relative to observed target-language data. Consequently, frequent target-language lemmas can override the Finnish prior, while rare or unattested lemmas can still benefit from the much larger Finnish corpus.

Morphological frequencies are estimated entirely from the target-language corpus. After finite-state analysis, counts are collected for morphological feature combinations such as \textit{Sg+Nom}, \textit{Pl+Nom}.

Because the corpora are small, many valid forms are not observed. Direct relative frequencies would therefore assign probability zero to forms that are grammatically possible but happen not to occur in the available texts. To avoid this, the observed counts are smoothed. An unobserved form receives probability from the broader distribution of morphological categories for the same part of speech, or, where available, for the same inflectional class.

Where sufficient evidence exists for an individual lemma, its own morphological distribution is used. For rare lemmas, the estimate is gradually backed off toward the general morphological distribution of the corresponding part of speech. Thus, a frequently occurring noun can develop a lemma-specific case and number distribution, whereas an unattested noun inherits the general noun inflection distribution.

Only analyses output by the FSTs are considered. If a lemma cannot occur in some morphological category, probability mass is redistributed among its legal forms. This prevents lemmas with restricted paradigms from being penalized simply because they have fewer possible inflected forms.

A surface form may have more than one valid morphological analysis. During corpus counting, such a token should not contribute a full count to every possible analysis, since this would artificially increase the size of the corpus. In the initial implementation, the contribution of an ambiguous token is divided equally among its analyses. Thus, a token with two analyses contributes $0.5$ to each.

At prediction time, the situation is reversed: if several analyses generate the same surface word, their probabilities are added together. The resulting probability therefore represents the probability of the surface form irrespective of which morphological analysis produced it.

The final probability of a surface word combines the target-language part-of-speech distribution, the Finnish-informed lemma probability, and the target-language morphological probability. For efficient ranking in the predictive keyboard, probabilities can equivalently be represented as negative log-probabilities,

\[
\mathrm{cost}(w) = -\log P(w),
\]

so that more probable words receive lower costs.

This approach has two main advantages for severely under-resourced languages. First, the very large Finnish corpus contributes information about the relative commonness of lexical meanings without numerically overwhelming the much smaller target-language corpus. Second, morphological behaviour is learned from the target language itself rather than transferred from Finnish. The model therefore makes use of large-scale lexical evidence while preserving language-specific inflectional patterns.

For deployment in the predictive keyboard, the resulting probabilities are converted into the integer weight representation required by HeliBoard. This transformation preserves the relative ranking of the estimated word-form probabilities while adapting them to the input format expected by the keyboard.

\section{Conclusions}

In this work, we presented open-source FUTO Keyboard layouts and predictive text dictionaries for Erzya, Moksha, Komi-Zyrian, Livonian, and Skolt Sami. By converting existing Divvun Android layouts, we were able to reuse established keyboard designs rather than create new ones from scratch. We then extended these layouts with weighted predictive dictionaries compatible with the AOSP dictionary format used by FUTO Keyboard.

The main challenge was the lack of large corpora for the target languages. To address this, we combined lexical frequency information transferred from Finnish through bilingual dictionaries with morphological frequency information extracted from small target-language corpora using finite-state transducers. This provides a practical compromise between purely rule-based generation and data-intensive statistical prediction.

The resulting workflow is simple, reproducible and applicable beyond the five languages considered here. More broadly, it shows how existing rule-based resources can be combined to improve everyday digital support for low-resource and endangered languages. Our code is open source and publicly available.



\bibliography{custom}

\end{document}